\documentclass[journal]{IEEEtran}

\usepackage{booktabs}
\usepackage{multirow}
\usepackage{color}
\usepackage{amssymb}
\usepackage{balance}
\usepackage{bm}
\usepackage{cite}

\usepackage[table, xcdraw]{xcolor}
\usepackage{makecell}

\ifCLASSINFOpdf
\usepackage[pdftex]{graphicx}
\graphicspath{{. . /pdf/}{. . /jpeg/}}
\DeclareGraphicsExtensions{. pdf, . jpeg, . png, . jpg}
\else
\fi
\usepackage{amsmath}
\usepackage{array}
\begin{document}
	\title{
	Novel Intensity Mapping Functions:
	\\Weighted Histogram Averaging
}
\author{
	Yilun~Xu, Zhengguo~Li,~\IEEEmembership{Fellow,~IEEE}, Weihai~Chen$^{\ast}$,~\IEEEmembership{Member,~IEEE,}
	and Changyun~Wen,~\IEEEmembership{Fellow,~IEEE}
	\thanks{$\ast$ Corresponding author: Weihai Chen. }
	\thanks{Yilun Xu and Weihai Chen are with School of Automation Science and Electrical Engineering, Beihang University, Beijing 100191, China. (e-mail: yilunxu\_buaa@163.com,  and whchen@buaa.edu.cn). }
	\thanks{Zhengguo Li is with SRO Department,  Institute for Infocomm Research,  1
		Fusionopolis Way,  Singapore (email: ezgli@i2r.a-star.edu.sg). }
	\thanks{Changyun Wen is with the School of Electrical and Electronics Engineering,
		Nanyang Technological University,  Singapore (email: ecywen@ntu.edu.sg). }
}
\maketitle

\begin{abstract}
	It is challenging to align the brightness distribution of the images with different exposures due to possible color distortion and loss of details in the brightest and darkest regions of input images.  In this paper,  a novel intensity mapping algorithm is first proposed by introducing a new concept of weighted histogram averaging (WHA).  The proposed WHA algorithm leverages the correspondence between the histogram bins of two images which are built up by using the non-decreasing property of the intensity mapping functions (IMFs). Extensive experiments indicate that the proposed WHA algorithm significantly surpasses the related state-of-the-art intensity mapping methods. 
\end{abstract}


\IEEEpeerreviewmaketitle

\section{Introduction}
\label{sec:intro}

Many color matching algorithms have been proposed in a dozen years \cite{CMF-CHM2003, CMF-GC2009, CMF-TIP2017, CMF-HHM2020} because they have broad application in many fields such as panorama stitching,  image fusion,  video intensity mapping,  etc \cite{CMF-re2016, CMF-re2010}.  Intensity mapping algorithms can be divided into model-based parametric algorithms and statistics-based non-parametric algorithms \cite{CMF-re2010}.  The model-based parametric algorithms assume that the intensity mapping from the original image to the target image follows a known mapping model.  The statistics-based non-parametric algorithms do not build an explicit mapping model but estimate the look-up table from the relevant statistics of the original image and target image,  and then use the look-up table as IMFs to calibrate the image.

In this paper, a novel HDR stitching algorithm is proposed by introducing a histogram bins-based intensity mapping algorithm\footnote{Intensity mapping algorithms can also be referred to as color mapping or color correction algorithms.}. The proposed intensity mapping algorithm includes estimation and correction of intensity mapping functions (IMFs).  The proposed intensity mapping method is inspired by two methods in \cite{CMF-GC2009,CMF-CHM2003}. The method in \cite{CMF-GC2009} first sets up a pixel level correspondence between the two differently exposed images, and then computes the mapped value through averaging over all the corresponding pixels in the other image. The method in \cite{CMF-GC2009} is very accurate if there is neither camera movement nor moving objects in the two images. However, the pixel-level correspondence is very sensitive to the camera movement and moving objects. The method in \cite{CMF-CHM2003} shows that the  histogram-bin-level correspondence is robust to the camera movement and moving objects even though the accuracy of the method in \cite{CMF-CHM2003} is an issue. Inspired by the above two methods, a novel histogram-bin-level correspondence is first built up between the two differently exposed images, i.e., \emph{each bin in the histogram of one image corresponds to one unique segment bins in the histogram of the other image}, by using the non-decreasing property of the IMFs. The mapped value is defined as the weighted average of intensities in the matched segment rather than the intensity of a single bin in the matched segment as in \cite{CMF-CHM2003}. The proposed method is thus termed weighted histogram averaging (WHA). Clearly, the proposed WHA algorithm is totally different from the methods in \cite{CMF-CHM2003, CMF-GC2009} even though it is inspired by them. Since the overlapping area of the sub-image does not necessarily contain all the pixel values in the dynamic range,  the calculated IMFs will contain some empty values. To make IMFs cover the whole dynamic range, linear interpolation and extrapolation are used to fill in the empty values after the IMFs are estimated. The proposed WHA algorithm is more accurate than the algorithm in \cite{CMF-CHM2003} and is more robust than the algorithm in \cite{CMF-GC2009} with respect to the camera movement and moving objects. Experimental results validate the proposed algorithms. Overall, the contributions of this paper is: a novel IMF estimation algorithm which outperforms existing IMF estimation algorithms including \cite{CMF-CHM2003, CMF-GC2009}.

The rest of this paper is organized as follows. Details of the proposed algorithm are presented in Section \ref{algo}. Extensive experimental results are provided in Section \ref{experiment}. Lastly, concluding remarks are listed in Section \ref{conclusion}.

\section{The Proposed IMF Estimation Method}
\label{algo}
In this section,  a novel IMF estimation algorithm that inherits the advantages of \cite{CMF-CHM2003, CMF-GC2009} is introduced.

\subsection{Initial Estimation of IMF via WHA}

Let $Z_i$ and $Z_j$ be two differently exposed images.
The pixel's position index and intensity of $Z_i$ are denoted by $p$ and $z$, respectively, where $p=(p_x,p_y)$. Let $\Lambda_{i \to j}(z)$ be the IMF from the image $Z_i$ to the image $Z_j$, and the mapped image $Z_{i \to j}$ can be calculated as follows
\begin{align}
	Z_{i \to j}(p) = \Lambda_{i \to j}(Z_i(p))
\end{align}
The position index set of pixels with the intensity value $z$ in the image $Z_i$, $\Omega_i(z)$ is defined as
\begin{align}
	\Omega_i(z)=\{p|Z_i(p)=z\}.
\end{align}

The IMF $\Lambda_{i \to j}^1(z)$ is estimated by using the cumulative histograms of the images $Z_i$ and $Z_j$ in \cite{CMF-CHM2003}.  Let $H_i(z)$ be the cardinality of the set $\Omega_i(z)$, and it represents the value of $z$th histogram bin of image $Z_i$. The width of the histogram bins defaults to 1. The cumulative histogram $C_i(z)$ of the image $Z_i$ is then created by
\begin{align}
	C_i(z) = \sum_{k=0}^{z} H_i(k).
\end{align}

This new method uses the histogram bin level correspondence  rather than the pixel level correspondence of the two differently exposed images as in \cite{CMF-GC2009}.  The correspondence among the histogram bins of the images $Z_i$ and $Z_j$ can be built up by using the non-decreasing property of the IMFs as follows:

\emph{Consider two pixels $Z_i(p_x,p_y)$ and $Z_i(p_x',p_y')$ which satisfy $Z_i(p_x,p_y)>Z_i(p_x',p_y')$. Suppose that the value of histogram bins in the image $Z_j$ corresponding to $Z_i(p_x,p_y)$ and $Z_i(p_x',p_y')$ are $H_j(z)$ and $H_j(z')$,  respectively.  $z$ is not smaller than $z'$.}

Since scene motion does not significantly change the distribution of image histograms, the above correspondence among the histogram bins is also true for the images $Z_i$ and $Z_j$  with camera movement and/or moving objects. The correspondence will be utilized to find all the sub-bins (or bins) in the image $Z_j$ corresponding to the bin $H_i(z)$. A non-decreasing mapping function $\psi_{i \to j}(z)$ is defined as
\begin{align}
	\label{CH:interval}
	C_j(\psi_{i \to j}(z)-1)< C_i(z) \leq C_j(\psi_{i \to j}(z)),
\end{align}
and $\psi_{i\to j}(-1)$ is predefined as 0.

According to the definitions of $C_i(z)$ and $C_j(z)$,  $H_j(\psi_{i \to j}(z-1))$ and $H_j(\psi_{i \to j}(z))$ are the first and last bins corresponding to the bin $H_i(z)$,  respectively. Let the sizes of sub-bins (or bins) in the image $Z_j$ corresponding to the bin $H_i(z)$be denoted by $\hat{H}_{i \to j}^z(k)$,  and it is defined as in the following two cases:

{\it Case 1}: $\psi_{i \to j}(z-1)<\psi_{i \to j}(z)$.  $\hat{H}_{i \to j}^z(k)$ is defined as
\begin{align}
	\resizebox{0.95\width}{!}{$
		\label{eq:calculation}
		\hat{H}_{i \to j}^z(k)=\left\{\begin{array}{ll}
			C_j(k) - C_i(z-1),  &\mbox{if~} k=\psi_{i \to j}(z-1)\\
			C_i(z)-C_j(k-1),  &\mbox{if~}k=\psi_{i \to j}(z)\\
			H_j(k),  &\mbox{otherwise}
		\end{array}
		\right.
		$}
\end{align}

{\it Case 2}: $\psi_{i \to j}(z-1)=\psi_{i \to j}(z)$.  $\hat{H}_{i \to j}^z(k)$ is defined as
\begin{align}
	\label{eq:calculation2}
	\hat{H}_{i \to j}^z(k)=	H_i(z).
\end{align}

It can be easily verified that
\begin{align}
	\label{CH:interval1}
	\sum_{k=\psi_{i \to j}(z-1)}^{\psi_{i \to j}(z)}\hat{H}_{i \to j}^z(k)=H_i(z).
\end{align}

By using the correspondence among the histogram bins,  the proposed IMF $\Lambda_{i \to j}(z)$ is then defined as
\begin{align}
	\label{eq:CHCH}
	\Lambda_{i \to j}(z)= \sum_{k=\psi_{i \to j}(z-1)}^{\psi_{i \to j}(z)}\frac{\hat{H}_{i \to j}^z(k)}{H_i(z)}k,  \ H_i(z) \neq0.
\end{align}

It can be shown from the equation (\ref{eq:CHCH}) that the proposed IMF $\Lambda_{i \to j}(z)$ is a weighted average of all the sub-bins (or bins) in the image $Z_j$ corresponding to the bins $H_i(z)$. Thus, the proposed IMF estimation algorithm is called  the WHA. The matlab code of the proposed WHA is give in the appendix. The definition of the function $\psi_{i \to j}(z)$ in the equation (\ref{CH:interval}) plays a crucial role in the proposed WHA. It builds up an elegant histogram-bin-level correspondence between the two differently exposed images. Unlike the pixel-level correspondence in \cite{CMF-GC2009},  the histogram-bin-level correspondence is robust to both the camera movement and the moving objects.

\subsection{Interpolation and Extrapolation of the Proposed IMF}

Since the image $Z_i$ does not necessarily include pixels in all the dynamic range,  there is often a problem of empty bins.  When $H_i(z) =0$,  the function $\Lambda_{i \to j}(z)$ can be calculated by using two neighboring $\Lambda_{i \to j}(z_1)$ and $\Lambda_{i \to j}(z_2)$,  and the formula is given as follows:
\begin{align}
	\Lambda_{i \to j}(z)=\frac{(z-z_1)(\Lambda_{i \to j}(z_1)-\Lambda_{i \to j}(z_2))}{z_1-z_2} + \Lambda_{i \to j}(z_1),
\end{align}
where the $z_1$ and $z_2$ can be estimated from
\begin{align}
	\min\quad & \{|z-z_1|+|z-z_2|\} \notag\\
	\mbox{s. t. }\quad
	&\lambda (z-z_1)(z-z_2)<0, \\
	&H_i(z_1)H_i(z_2) \neq 0 \notag.
\end{align}
Here,  $\lambda$ is set as 1 for the interpolation,  and -1 for the extrapolation.

\section{Experiment Results}
\label{experiment}

Extensive experimental results are provided to verified the proposal algorithms.  Readers are invited to view to electronic version of figures and zoom in them so as to better appreciate differences among all images.

\subsection{Implementation Details}

\subsubsection{Dataset Description}
Experiments are conducted on the VETHDR-Nikon dataset. The VETHDR-Nikon dataset is from \cite{dataset-chaobing},  collected by a Nikon 7200 camera.  It consists of 495 pairs of images in 8-bit JPEG files format.  Each pair consists of two full-resolution images,  with long and short exposure times,  respectively.  The exposure time ratio and ISO are fixed as 8 and 800,  during shooting respectively.  All the images are resized to $1000 \times 1600$.  Camera shaking,  object movement are strictly controlled to ensure that only the illumination is changed.

\subsubsection{IMFs Comparison Description}
\begin{figure}[htbp]
	\setlength{\abovecaptionskip}{0pt}
	\setlength{\belowcaptionskip}{0pt}
	\begin{center}
		\includegraphics[width=1.0\linewidth]{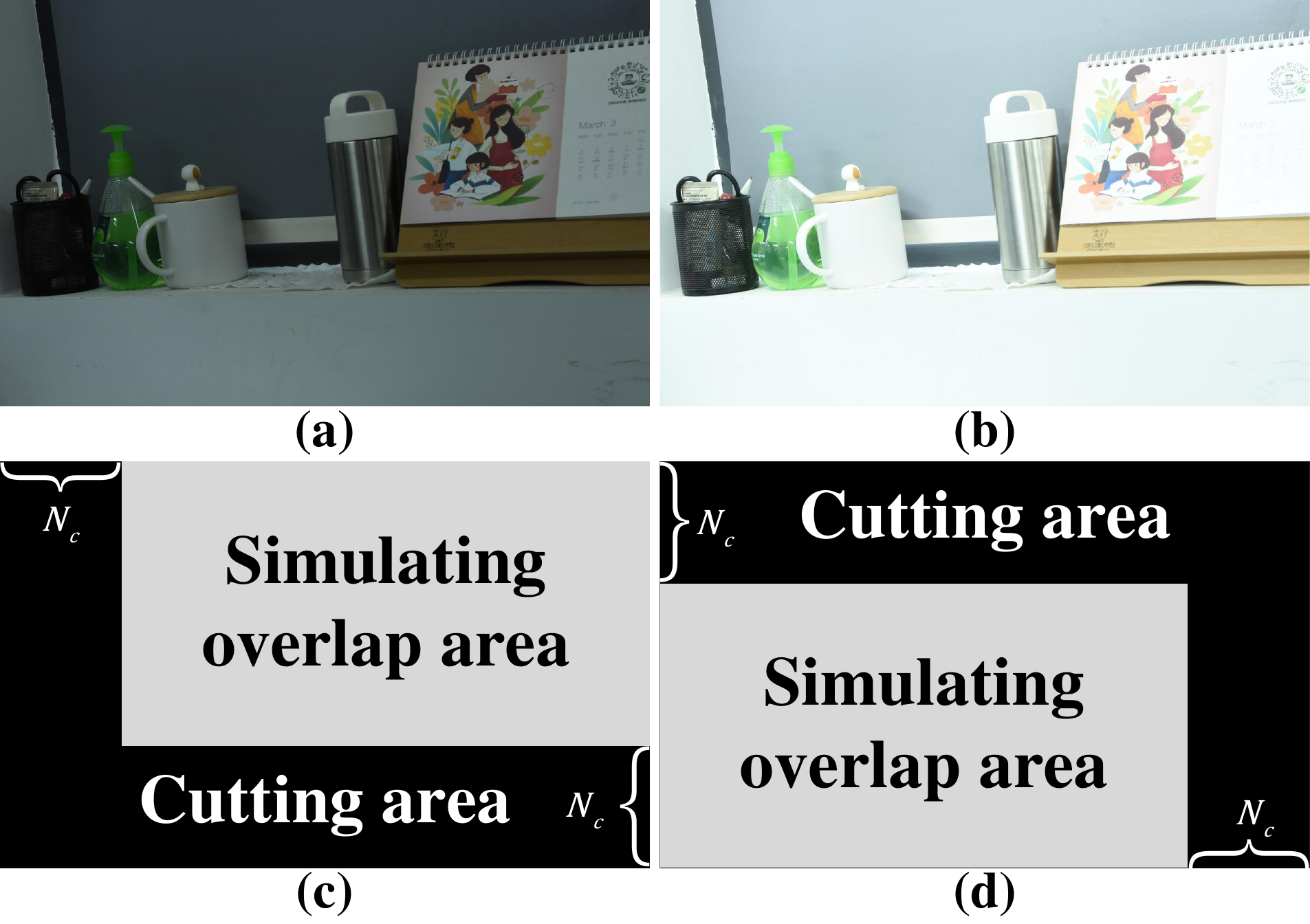}
	\end{center}
	\caption{An example of simulating overlapping areas. (a) and (b) are the input image pairs, (c) and (d) show the way to simulate the overlapping area of (a) and (b), respectively.}
	\label{fig:overlap}
\end{figure}

In the VETHDR-Nikon dataset \cite{dataset-chaobing},  the overlapping areas in each pair of images are well-aligned.  However,  in actual situations,  since the shooting angles of the two images to be stitched are different,  the overlapping areas are always not well-aligned.  In order to simulate the real situation of the overlapping area,  Fig.  \ref{fig:overlap} shows how to process the input image.  Specifically,  the $N_c$ columns of pixels on the left and $N_c$ rows of pixels on the bottom of the image (a) are cut off,  the $N_c$ columns of pixels on the right and $N_c$ rows of pixels on the top of the image (b) are also cut off.  The remaining areas are used as the simulated overlap area. The IMFs are first estimated by the simulated overlap areas and then used to correct the brightness of image (a) or (b).

\begin{figure*}[htbp]
	\setlength{\abovecaptionskip}{0pt}
	\setlength{\belowcaptionskip}{0pt}
	\begin{center}
		\includegraphics[width=1.0\linewidth]{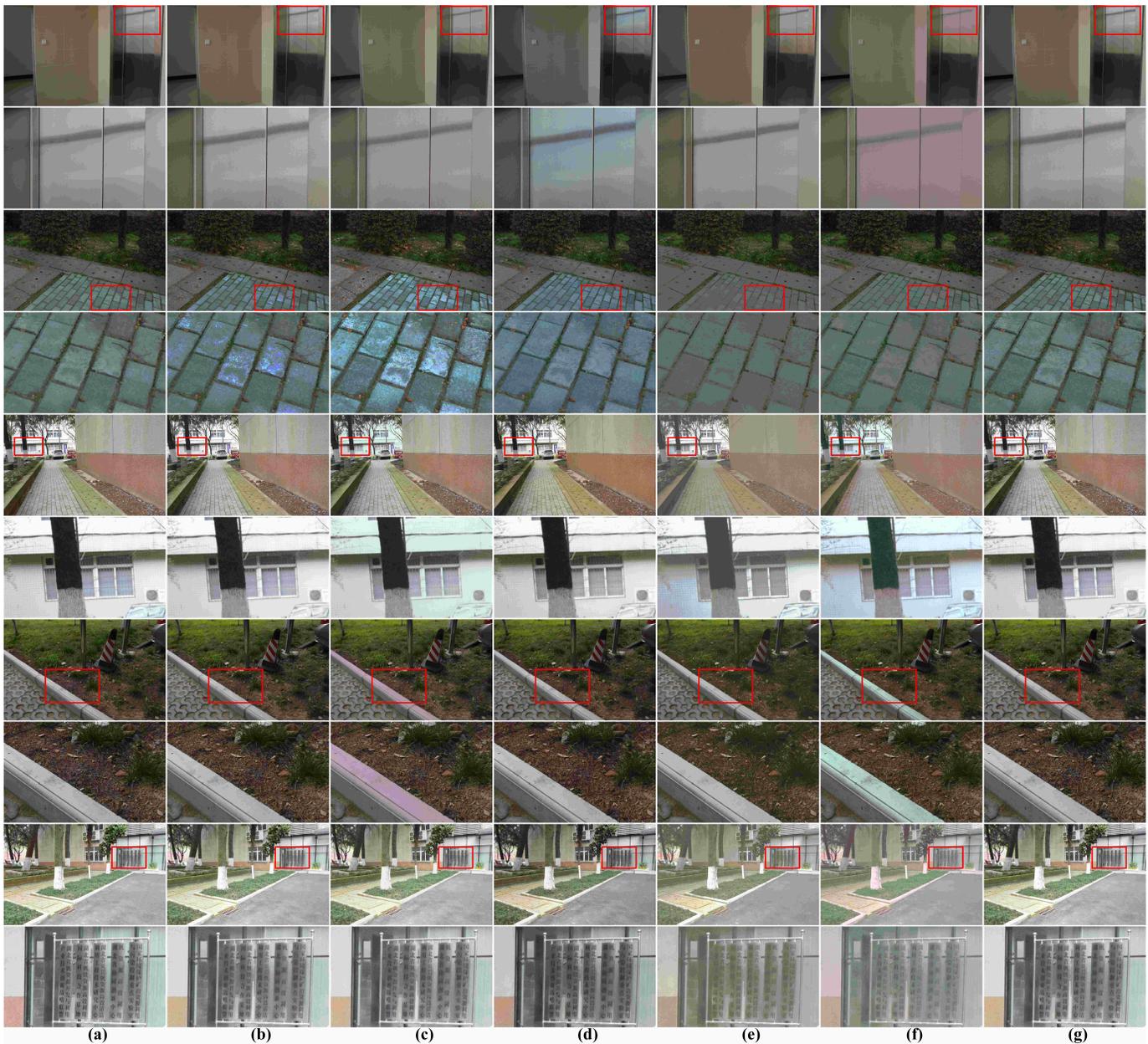}
	\end{center}
	\caption{Qualitative comparison among the existing IMFs' algorithms and the proposed WHA. (a) Original Images, (b) Ground truth, (c) the CHM \cite{CMF-CHM2003}, (d) the HHM \cite{CMF-HHM2020}, (e) the PCRF \cite{CMF-PCRF2018}, (f) the GPS \cite{CMF-GPS2017}, (g) the proposed WHA. The inputs are  from the VETHDR-Nikon dataset \cite{dataset-chaobing}.}
	\label{fig:exp_IMFs}
\end{figure*}

Moreover,  in order to test the IMFs more comprehensively, each pair of images in the VETHDR-Nikon dataset will be tested twice.  Specifically,  use the dark image as the reference to correct the bright image,  and then use the bright image as the reference to correct the dark image.  So each IMF estimation algorithm will be tested 990 times on the VETHDR-Nikon dataset \cite{dataset-chaobing}.

In terms of quantitative comparison,  we choose PSNR,  SSIM,  FSIM,  iCID and running time as the evaluation metrics.  All results are tested on a laptop with Intel Core i7-9750H CPU 2. 59GHz,  32. 0 GB memory and Matlab R2019a installed.  The all evaluation metrics are explained as follows:
\begin{itemize}
	\item \textbf{PSNR}: Peak signal-to-noise ratio,  the higher score comes with the better performance.
	\item \textbf{SSIM}: Structural similarity \cite{SSIM},  the higher score comes with the better performance.
	\item \textbf{FSIM}: Feature similarity \cite{FSIM},  the higher score comes with the better performance.
	\item \textbf{iCID}: Improved color image difference \cite{iCID},  the lower score comes with the better performance.
	\item \textbf{Time}: The time it takes for the IMFs algorithm to map a image,  it does not include the time to read and save the image.
\end{itemize}

Table \ref{tab:Description} summarizes the IMF algorithms involved in the experiments.
\begin{table}[htbp]
	\setlength{\abovecaptionskip}{0pt}
	\setlength{\belowcaptionskip}{0pt}
	\centering
	\scriptsize
	\caption{a brief summary of IMFs algorithms.
		\label{tab:Description}
	}\rowcolors[]{2}{white!100}{gray!15}
	\resizebox{1.0\columnwidth}{!}{\begin{tabular}{c||l}
			\toprule[1pt]
			\textbf{Method} & \multicolumn{1}{c} {\textbf{Summary Description}}  \\
			\midrule
			CHM \cite{CMF-CHM2003} &  using  cumulative histogram\\
			GC \cite{CMF-GC2009} & using  geometric correspondence between pixels\\
			HHM \cite{CMF-HHM2020} & using hybrid histogram matching algorithm\\
			PCRF \cite{CMF-PCRF2018} &  using prior information about camera response function\\
			TG \cite{CMF-TG2014} &  using histogram simulated by truncated Gaussians\\
			AM \cite{CMF-AM2014} &  using affine map\\
			3MS \cite{CMF-3MS2013} &  using monotonic splines and three-channel correlation\\
			GPS \cite{CMF-GPS2017} &  using gradient preserving spline\\
			MV \cite{CMF-2001} &  using the mean and variance of the images\\
			\midrule
			WHA & using weighted histogram averaging algorithm\\
			\bottomrule[1pt]
	\end{tabular}}
\end{table}

\subsection{Comparison with Existing Intensity Mapping Algorithms}

The proposed WHA algorithm is compared with nine existing intensity mapping algorithms in \cite{CMF-HHM2020, CMF-PCRF2018, CMF-TG2014, CMF-AM2014, CMF-3MS2013, CMF-GPS2017, CMF-2001, CMF-GC2009, CMF-CHM2003} qualitatively and quantitatively on the VETHDR-Nikon dataset \cite{dataset-chaobing}.  Table \ref{tab:Description} summarizes all these algorithms.

$N_c$ is set to 10 in order to verify the accuracy and robustness of all the IMF estimation algorithms simultaneously.  The results of 5 evaluation metrics are reported in Table \ref{tab:exp_IMFs}.  With the second fastest running speed,  our method achieves the best performance in all the PSNR,  SSIM,  FSIM,  and iCID.
Both the proposed WHA and the CHM \cite{CMF-CHM2003} outperform the PCRF \cite{CMF-PCRF2018}. Clearly, the IMFs should be adaptive to the statistics between each pair of images to achieve \emph{instance adaption}.

\begin{table}[htbp]
	\setlength{\abovecaptionskip}{0pt}
	\setlength{\belowcaptionskip}{0pt}
	\centering
	\scriptsize
	\caption{Quantitative comparison among the existing intensity mapping algorithms. The best results are shown in bold, and the second-best results are shown in red.
		\label{tab:exp_IMFs}
	}\rowcolors[]{2}{white!20}{gray!15}
	\resizebox{1.0\columnwidth}{!}{\begin{tabular}{c||ccccccccc}
			\toprule[1pt]
			\textbf{Method} & \textbf{PSNR$\uparrow$} & \textbf{SSIM$\uparrow$} & \textbf{FSIM$\uparrow$} & \textbf{iCID(\%)$\downarrow$} & \textbf{Time(s)$\downarrow$}  \\
			\midrule
			HHM \cite{CMF-HHM2020} & 29.88 & 0.8977 & 0.9673 & 8.24  & 0.11  \\
			PCRF \cite{CMF-PCRF2018} & 30.69 & {\color[HTML]{FF0000} 0.9056} & 0.9715 & 7.67  & 1.79  \\
			TG \cite{CMF-TG2014} & 29.34 & 0.8897 & 0.9391 & 11.35  & 12.52  \\
			AM \cite{CMF-AM2014}& 25.68 & 0.8413 & 0.8882 & 18.49  & 0.54  \\
			3MS \cite{CMF-3MS2013}& 28.96 & 0.8857 & 0.9455 & 11.17  & 7.70  \\
			GPS \cite{CMF-GPS2017} & 30.96 & 0.8945 & 0.9643 & 9.36  & 151.09  \\
			MV \cite{CMF-2001} & 28.14  & 0.8964  & 0.9470  & 10.28  &  0.66  \\
			GC \cite{CMF-GC2009}& 28.17 & 0.8693 & 0.9385 & 11.98  & 1.77  \\
			CHM \cite{CMF-CHM2003} & {\color[HTML]{FF0000} 32.36} & 0.9029 & {\color[HTML]{FF0000} 0.9740} & {\color[HTML]{FF0000} 7.10} & \textbf{0.08}  \\
			\midrule
			WHA & \textbf{34.38} & \textbf{0.9153} & \textbf{0.9815} & \textbf{4.77}  &  {\color[HTML]{FF0000} 0.08}  \\
			\bottomrule[1pt]
	\end{tabular}}
	
\end{table}
For the qualitative comparison,  several representative methods were selected.  The visualization results and their detailed parts are shown in Fig.  \ref{fig:exp_IMFs}.  In the first row,  the color of the door is mapped incorrectly by the algorithms in \textbf{(d)},  \textbf{(e)} and \textbf{(f)}.  In the second row,  the luminance of images in \textbf{(d)} and \textbf{(f)} is different from the ground-truth,  and the color of images in \textbf{(c)},  \textbf{(e)} and \textbf{(f)} is abnormal.  In a nutshell,  the results in \textbf{(d)},  \textbf{(e)} and \textbf{(f)} are likely to have abnormal color and luminance,  the results in \textbf{(c)} sometimes have abnormal color when mapped from lighter images.  Different from these methods,  the proposed algorithm is robust to most scenes,  providing accurate results.  The overall comparison demonstrates the effectiveness of the proposed WHA algorithm.

\section{Conclusion Remarks and Discussion}
\label{conclusion}
In this paper,  a novel intensity mapping estimation algorithm is proposed by introducing a new concept of weighted histogram averaging (WHA),  and it is applied to study image stitching. Instead of only correcting the color and brightness difference as in the existing image stitching algorithms,  a set of differently exposed low dynamic range (LDR) panorama images is first synthesized and then fused via a multi-scale exposure fusion algorithm to produce the desired HDR panorama images.  Details in the brightest and darkest regions of the input images are preserved in the HDR panorama images much better than traditional LDR stitching algorithms.

\section*{Acknowledgment}
This work was supported by the National Natural Science Foundation of China under Grant No. U1909215, the Key Research and Development Program of Zhejiang Province under Grant No. 2021C03050, the Scientific Research Project of Agriculture and Social Development of Hangzhou under Grant No. 2020ZDSJ0881, and the National Natural Science Foundation of China under the research project 61620106012, 61573048.

\bibliographystyle{IEEEtran}
\bibliography{refs}	
\end{document}